\documentclass[10pt,twocolumn,letterpaper]{article}

\usepackage{cvpr}
\usepackage{times}
\usepackage{epsfig}
\usepackage{graphicx}
\usepackage{amsmath}
\usepackage{amssymb}
\usepackage{subfig}
\usepackage{multirow}

\usepackage{amsmath}
\usepackage{amssymb}
\usepackage{mathrsfs}
\usepackage{amsthm}
\usepackage{color}

\newcommand{\hide}[1]{}

\newcommand{\bd}{\mathbf{d}}

\newcommand{\bp}{\mathbf{p}}

\newcommand{\bx}{\mathbf{x}}

\newcommand{\bA}{\mathbf{A}}

\newcommand{\bC}{\mathbf{C}}
\newcommand{\bD}{\mathbf{D}}

\newcommand{\bI}{\mathbf{I}}

\newcommand{\bU}{\mathbf{U}}
\newcommand{\bV}{\mathbf{V}}
\newcommand{\bW}{\mathbf{W}}

\newcommand{\bLambda}{\boldsymbol{\Lambda}}

\newcommand{\nystrom}{Nystr\"{o}m }
\usepackage[pagebackref=true,breaklinks=true,letterpaper=true,colorlinks,bookmarks=false]{hyperref}

\cvprfinalcopy %

\def\cvprPaperID{1418} %

\ifcvprfinal\fi
\begin{document}

\title{Pooling-Invariant Image Feature Learning}

\author{Yangqing Jia\\
UC Berkeley\\
{\tt\small jiayq@eecs.berkeley.edu}
\and
Oriol Vinyals\\
UC Berkeley\\
{\tt\small vinyals@eecs.berkeley.edu}
\and
Trevor Darrell\\
UC Berkeley \& ICSI\\
{\tt\small trevor@eecs.berkeley.edu}
}

\maketitle

\begin{abstract}
Unsupervised dictionary learning has been a key component in state-of-the-art computer vision recognition architectures. While highly effective methods exist for patch-based dictionary learning, these methods may learn redundant features after the pooling stage in a given early vision architecture. In this paper, we offer a novel dictionary learning scheme to efficiently take into account the invariance of learned features after the spatial pooling stage.
The algorithm is built on simple clustering, and thus enjoys efficiency and scalability. We discuss the underlying mechanism that justifies the use of clustering algorithms, and empirically show that the algorithm finds better dictionaries than patch-based methods with the same dictionary size. 
\end{abstract}

\section{Introduction}

In the recent decade local patch-based, spatially pooled feature extraction pipelines have been shown to provide good image features for classification. Methods following such a pipeline usually start from densely extracted local image patches (either normalized raw pixel values or hand-crafted descriptors such as SIFT or HOG), and perform dictionary learning to obtain a dictionary of codes (filters). The patches are then encoded into an over-complete representation using various algorithms such as sparse coding \cite{Olshausen:1997uh,wang2010locality} and simple inner product with a non-linear post-processing \cite{coates2011icml,krizhevsky2012imagenet}. After encoding, spatial pooling with average or max operations are carried out to form a global image representation \cite{Yang:2009vb,Boureau:uq}. The encoding and pooling pipeline can be stacked to produce a final feature vector, which is then used to predict the labels for the images usually via a linear classifier.

There is an abundance of literature on single-layered networks for unsupervised feature encoding. Dictionary learning algorithms have been discussed to find a set of basis that reconstructs local image patches or descriptors well \cite{mairal2010online,coates2011icml}, and several encoding methods have been proposed to map the original data to a high-dimensional space that emphasizes certain properties, such as sparsity \cite{Olshausen:1997uh,Yang:2009vb,yang2010efficient} or locality \cite{wang2010locality}. 

A particularly interesting finding in the recent papers \cite{coates2010aistats, Rigamonti:2011uc, coates2011icml, saxe2011random} is that very simple patch-based dictionary learning algorithms like K-means or random selection, combined with feed-forward encoding methods with a naive nonlinearity, produces state-of-the-art performance on various datasets. Explanation of such phenomenon often focuses on the local image patch statistics, such as the frequency selectivity of random samples \cite{saxe2011random}.

A potential problem with such patch-based learning methods is that it may learn redundant features when we consider the pooling stage, as two codes that are uncorrelated may become highly correlated after pooling due to the introduction of spatial invariance. While using a larger dictionary almost always alleviates this problem, in practice we often want the dictionary to have a limited number of codes due to various reasons. First, feature computation has become the dominant factor in the state-of-the-art image classification pipelines, even with purely feed-forward methods (\eg, threshold encoding \cite{coates2011icml}) or speedup algorithms (\eg, LLC \cite{wang2010locality}). Second, reasonably sized dictionary helps to more easily learn further tasks that depends on the encoded features; this is especially true when we have more than one coding-pooling stage such as stacked deep networks, or when one applies more complex pooling stages such as second-order pooling \cite{carreirasemantic}, as a large encoding output would immediately drive up the number of parameters in the next layer.

\begin{figure*}
    \centering
    \includegraphics[width=0.9\textwidth]{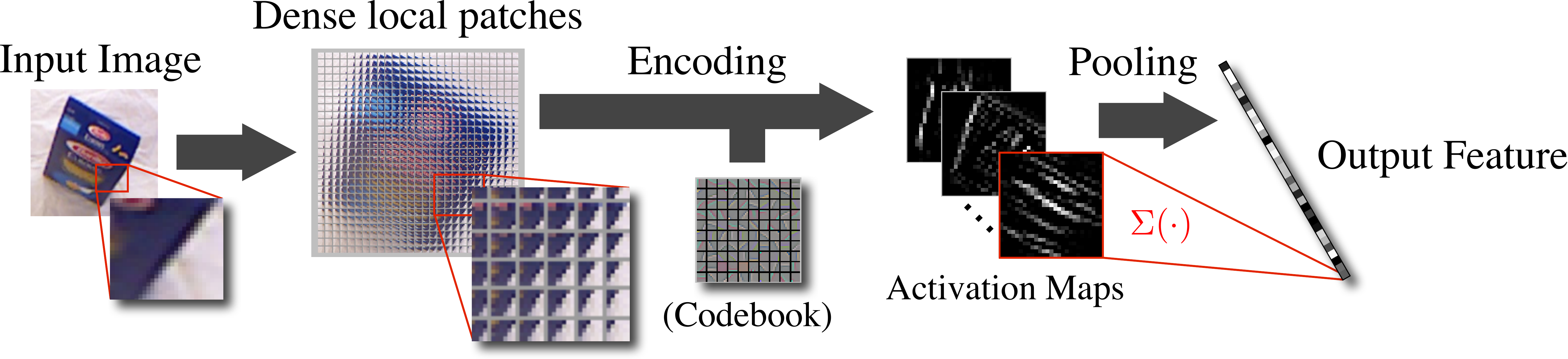}
    \caption{The feature extraction pipeline, composed of dense local patch extraction, encoding, and pooling. Illustrated is the average pooling over the whole image for simplicity, and in practice the pooling can be carried out over finer grids of the image as well as with different operations (such as max).}\label{fig:pipeline}
\end{figure*}

Thus, it would be beneficial to design a dictionary learning algorithm that takes pooling into consideration and learns a compact dictionary. Prior work on addressing such problem often resorts to convolutional approaches \cite{lee2009convolutional,zeiler2010deconvolutional}. These methods are usually able to find dictionaries that bear a better level of spatial invariance than patch-based K-means, but are often non-trivial to train when the dictionary size is large, since a convolution operation has to be carried out instead of simple inner products.

In our paper we present a new method that is analogous to the patch-based K-means method for dictionary learning, but takes into account the redundancy that may be introduced in the pooling stage. We show how a K-centroids clustering method applied on the covariance between candidate codes can efficiently learn pooling invariant representations. We also show how one can view dictionary learning as a matrix approximation problem, which finds the best approximation to an ``oracle'' dictionary (which is often very large). It turns out that under this perspective, the performance of various dictionary learning methods can be explained by the recent findings in \nystrom subsampling \cite{kumar2012sampling,zhang2008improved}.

We will first review the feature extraction pipeline and the effect of pooling on the learned dictionary in Section \ref{sec:nystrom}, and then describe the proposed two-stage dictionary learning method in Section \ref{sec:algorithm}. The effectiveness of this simple yet effective dictionary learning algorithms on the standard CIFAR-10 and STL benchmark datasets, as well as the fine-grained classification task, in which we show that feature learning plays an important role.

\section{Background}\label{sec:nystrom}
We illustrate the feature extraction pipeline that is composed of encoding dense local patches and pooling encoded features in Figure \ref{fig:pipeline}. Specifically, starting with an input image $\bI$, we formally define the encoding and pooling stages as follows.

\paragraph{(1) Coding.} In the coding step, we extract local image patches\footnote{Although we use the term ``patches'' throughout the paper, the pipeline works with local image descriptors, such as SIFT, as well.}, and encode each patch to $K$ activation values based on a dictionary of size $K$ (learned via a separate dictionary learning step). These activations are typically binary (in the case of vector quantization) or continuous (in the case of e.g.\ sparse coding), and it is generally believed that having an over-complete ($K >$ the dimension of patches) dictionary while keeping the activations sparse helps classification, especially when linear classifiers are used in the later steps.

We will mainly focus on what we call the decoupled encoding methods, in which the activation of one code does not rely on other codes, such as threshold encoding \cite{coates2011icml}, which computes the inner product between $\bp$ and each code, with a fixed threshold parameter $\alpha$: $f_{k}(\bx) = \max\{0, \bd_k^\top\bp - \alpha\}$. Such methods have been increasingly popular mainly for their efficiency over coupled encoding methods such as sparse coding, for which a joint optimization needs to be carried out. Their employment in several deep models (e.g. \cite{krizhevsky2012imagenet}) also suggests that such simple non-linearity may suffice to learn a good classifier in the later stages.

{\bfseries Learning the dictionary:} Recently, it has been found that relatively simple dictionary learning and encoding approaches lead to surprisingly good performances \cite{coates2010aistats,saxe2011random}. For example, to learn a dictionary $\mathcal{D} =\{\bd_1,\bd_2,\cdots,\bd_K\}$ of size $K$ from randomly sampled patches $\mathcal{P} = \{\bp_1,\bp_2,\cdots,\bp_N\}$ each reshaped as a vector of pixel values, one could simply adopt the K-means algorithm, which aims to minimize the squared distance between each patch and its nearest code: $\min_{\bD} \sum_{i=1}^{N}\min_{j}\|\bp_i - \bd_j\|_2^2$. We refer to \cite{coates2010aistats} for a detailed comparison about different dictionary learning and encoding algorithms.

\begin{figure*}
    \centerline{%
        (a)~\includegraphics[width=0.45\linewidth]{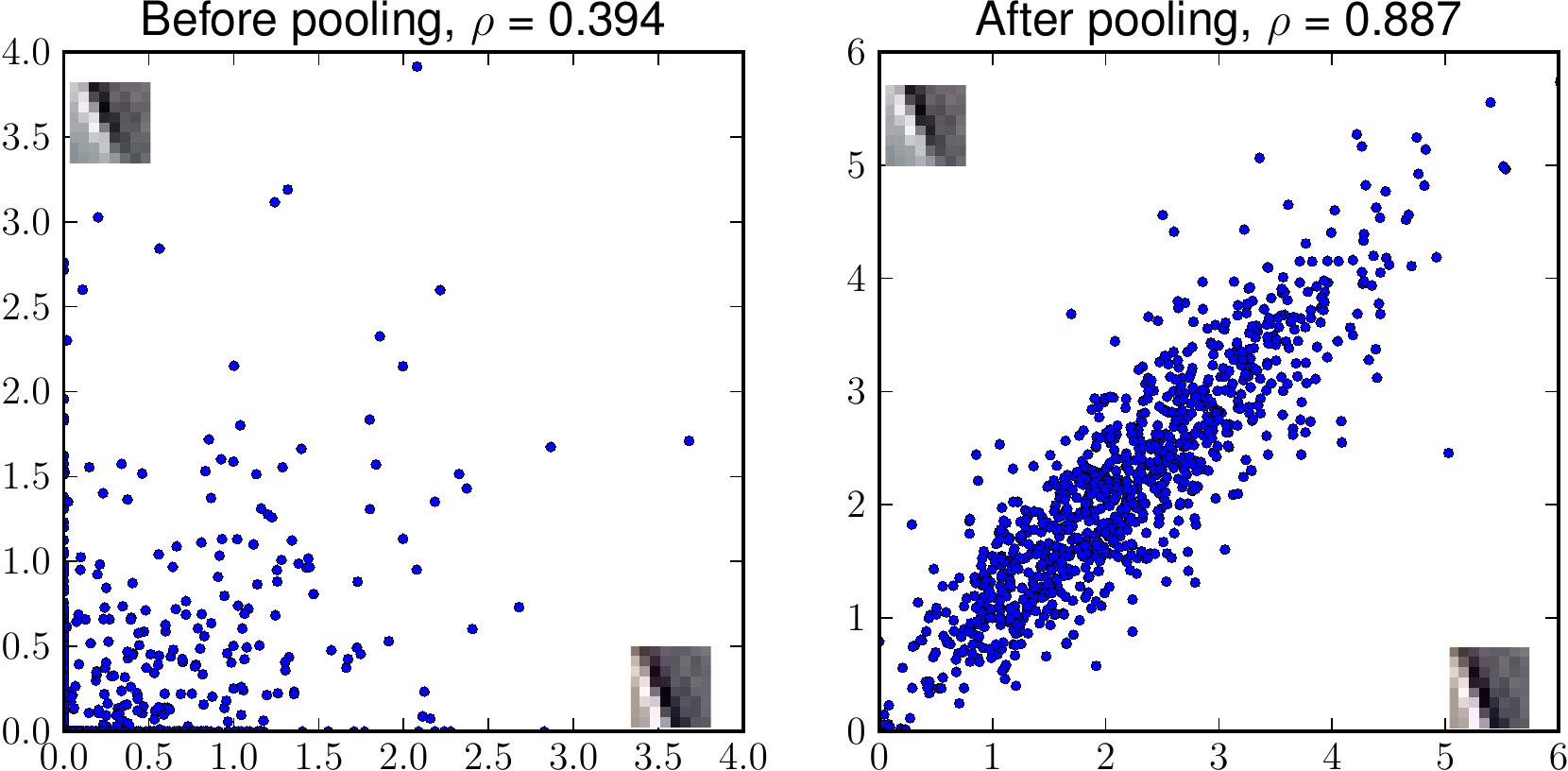}%
        ~~~~~~~%
        (b)~\includegraphics[width=0.45\linewidth]{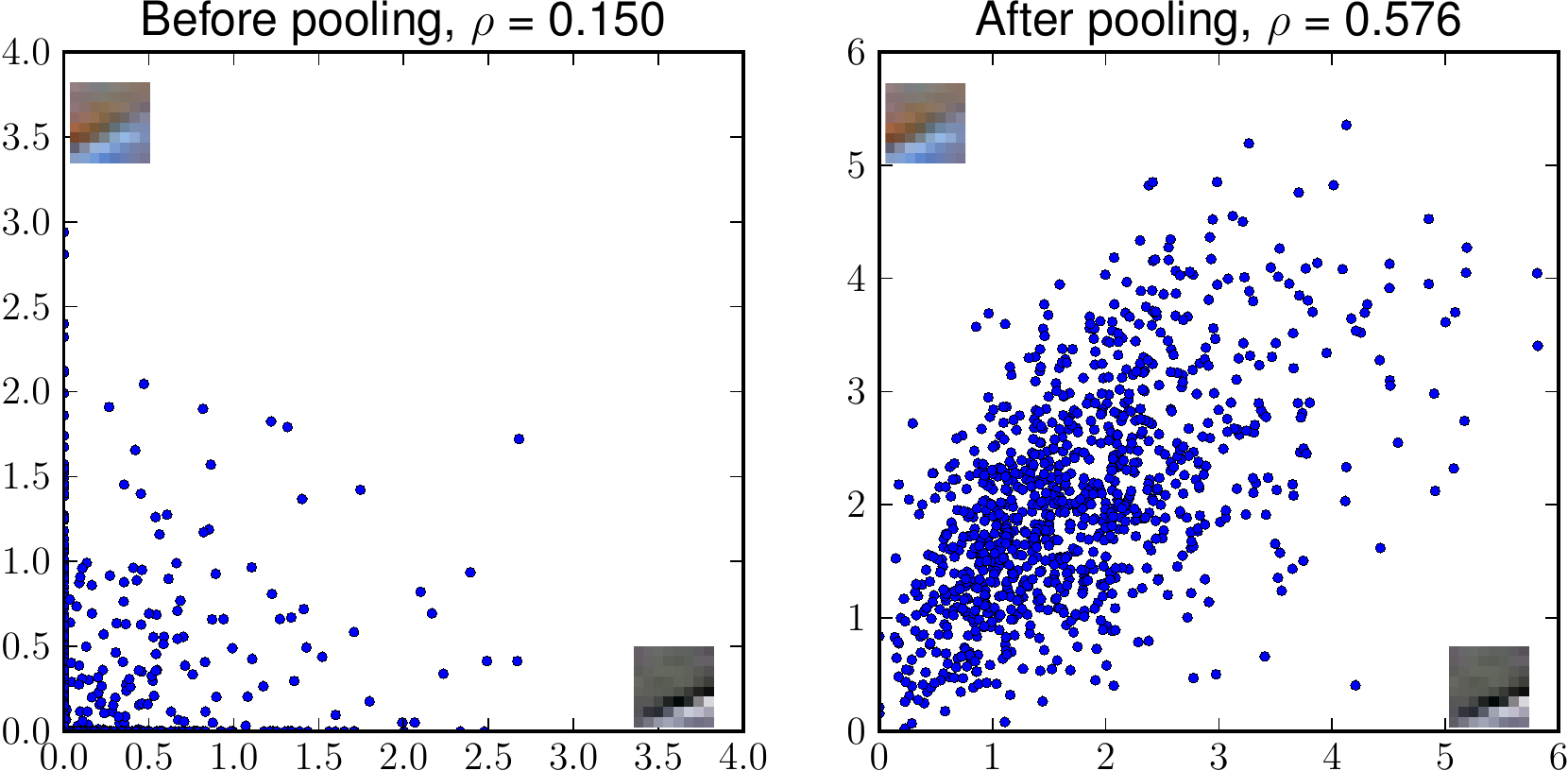}%
    }
    \caption{Two codes learned from a patch-based K-means algorithm that produce lowly correlated patch-based responses, but highly correlated responses after pooling. Notice that such phenomenon does not only exist between codes with translational difference (as in (a)), but also between other appearances (such as color in (b)).}\label{fig:feature_correlation}
\end{figure*}

\paragraph{(2) Pooling.} Since the coding result are highly over-complete and highly redundant, the pooling layer aggregates the activations over a spatial region of the image to obtain a $K$ dimensional vector $\bx$. Each dimension of the pooled feature $\bx_i$ is obtained by taking the activations of the corresponding code in the given spatial region (also called receptive field in the literature), and performing a predefined operator (usually average or max) on the set of activations.

Figure \ref{fig:pipeline} shows an example when average pooling is carried out over the whole image. In practice we may define multiple spatial regions per image (such as a regular grid), and the global representation for the image will then be a vector of size $K$ times the number of spatial regions.

Since the feature extraction for image classification often involves the spatial pooling stage, the patch-level dictionary learning may not find good dictionaries that produce most informative pooled outputs. In fact, one would reasonably argue that it doesn't, one immediate reason being that patch-based dictionary learning algorithms often yield similar Gabor filters with small translations. Such filters, when pooled over a certain spatial region, produce highly correlated responses and lead to redundancy in the feature representation. Figure \ref{fig:feature_correlation} shows two such examples, where filters produce uncorrelated patch-based responses but highly correlated pooled responses.

Convolutional approaches \cite{lee2009convolutional,zeiler2010deconvolutional} are usually able to find dictionaries that are more spatially invariant than patch-based K-means, but learning may not scale as well as simple clustering algorithms, especially with hundreds or thousands of codes. In addition, convolutional approaches may still not solve the problem of inter-code invariance: for example, the response of a colored edge filter might have high correlation with that of a gray scale edge filter, and such correlation could not be modeled by spatial invariance.

\section{Pooling-Invariant Dictionary Learning}\label{sec:algorithm}
We are interested in designing a simple yet effective dictionary learning algorithm that takes into consideration the pooling stage of the feature extraction pipeline, and that models the general invariances among the pooled features. Observing the effectiveness of clustering methods in dictionary learning, we propose to learn a final dictionary of size $K$ in two stages: first, we adopt the patch-based K-means algorithm to learn a more over-complete starting dictionary of size $M$ ($M>K$); we then perform encoding and pooling using the dictionary, learn the final, smaller dictionary of size $K$ from the statistics of the $M$ pooled features. 

The motivation of such idea is that K-means is a highly parallelizable algorithm that could be scaled up by simply sharding the data, allowing us to have an efficient algorithm for dictionary learning. Using a starting dictionary allows us to preserve most information on the patch-level, and the second step prunes away the redundancy due to pooling. Note that the large dictionary is only used during the feature learning time - after this, for each input image, we only need to encode local patches with the selected, relatively smaller dictionary, not any more expensive than existing feature extraction methods.

\subsection{Feature Selection with Affinity Propagation}
The first step of our algorithm is identical to the patch-based K-means algorithm with a starting dictionary size $M$. After this, we can sample a set of image super-patches of the same as the pooling regions, and obtain the $M$ dimensional pooled features from them. Randomly sampling a large number of pooled features in this way allows us to analyze the pairwise similarities between the codes in the starting dictionary in a post-pooling fashion. We would then like to find a $K$-dimensional subspace that best represents the $M$ pooled features. Specifically, the similarity between two pooled dimensions (which correspond to two codes in the starting dictionary) $i$ and code $j$ as
\begin{equation}
    s(i,j) = \frac{2C_{ij}}{\sqrt{C_{ii}C_{jj}}} - 2
\end{equation}
where $\bC$ is the covariance matrix computed from the random sample of pooled features. We note that this is equivalent to the negative Euclidean distance between the coded output $i$ and the coded output $j$ when the outputs are normalized to have zero mean and standard deviation 1. We then use affinity propagation \cite{frey2007clustering}, which is a version of the K-centroids algorithm, to select centroids from existing features. Intuitively, codes that produce redundant pooled output (such as translated versions of the same code) would have high similarity between them, and only one exemplar would be chosen by the algorithm.

Specifically, affinity propagation finds centroids from a set of candidates where pairwise similarity $s(i,j)$ ($1 \leq i, j \leq M$) can be computed. It iteratively updates two terms, the ``responsibility'' $r(i,j)$ and the ``availability'' $a(i,j)$ via a message passing method following such rules \cite{frey2007clustering}:
\begin{align}
    r(i,k) & \leftarrow s(i,k) - \max_{k'\neq k}\{a(i,k') + s(i,k')\}\\
    a(i,k) & \leftarrow \min\{0, r(k,k) + \sum\nolimits_{i'\notin\{i,k\}}\max\{0, r(i',k)\}\}\nonumber\\
           & \phantom{\leftarrow } (\text{if } i \neq k)\\
    a(k,k) & \leftarrow \sum\nolimits_{i'\neq k} \max\{0, r(i', k)\}
\end{align}
Upon convergence, the centroid that represents any candidate $i$ is given by $\arg\max_{k} (a(i,k) + r(i,k))$, and the set of centroids $\mathcal{S}$ is obtained by
\begin{equation}
    \mathcal{S} = \{k | \exists i,k \text{ s.t. } k = \arg\max_{k'} (a(i,k') + r(i,k'))\}
\end{equation}
And we refer to \cite{frey2007clustering} for details about the nature of such message passing algorithms.

\begin{figure*}
    \centering
        \includegraphics[height=0.2\linewidth]{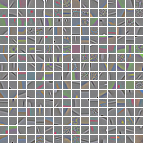}\qquad\qquad
        \includegraphics[height=0.2\textwidth]{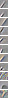}
        \includegraphics[height=0.2\textwidth]{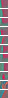}
        \includegraphics[height=0.2\textwidth]{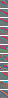}
        \includegraphics[height=0.2\textwidth]{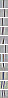}
        \includegraphics[height=0.2\textwidth]{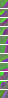}
        \includegraphics[height=0.2\textwidth]{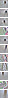}
        \includegraphics[height=0.2\textwidth]{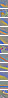}
        \includegraphics[height=0.2\textwidth]{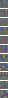}
        \includegraphics[height=0.2\textwidth]{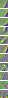}
        \includegraphics[height=0.2\textwidth]{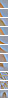}
        \includegraphics[height=0.2\textwidth]{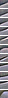}
        \includegraphics[height=0.2\textwidth]{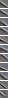}
        \includegraphics[height=0.2\textwidth]{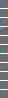}
        \includegraphics[height=0.2\textwidth]{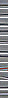}
        \includegraphics[height=0.2\textwidth]{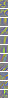}
        \includegraphics[height=0.2\textwidth]{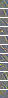}
        \includegraphics[height=0.2\textwidth]{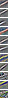}
    \caption{Visualization of the learned codes. Left: the selected subset of 256 centroids from an original set of 3200 codes. Right: The similarity between each centroid and the other codes in its cluster. For each column, the first code is the selected centroid, and the remaining codes are in the same cluster represented by it. Notice that while translational invariance is the most dominant factor, our algorithm does find invariances beyond that (e.g., notice the different colors on the last column). Best viewed in color.}\label{fig:centroidcodes}
\end{figure*}

\subsection{Visualization of Selected Filters}\label{subsec:visualization}
To visually show what codes are selected by affinity propagation, we applied our approach to the CIFAR-10 dataset by first training an over-complete dictionary of 3200 codes, and then performing affinity propagation on the 3200-dimensional pooled features to obtain 256 centroids, which we visualize in Figure \ref{fig:centroidcodes}. Translational invariance appears to be the most dominant factor, as many clusters contain translated versions of the same Gabor like code, especially for gray scale codes. On the other hand, clusters capture more than translation: clusters such as column 5 focus on finding the contrasting colors more than finding edges of exactly the same angle, and clusters such as the last column finds invariant edges of varied color. We note that the selected codes are not necessarily centered (which is the case for convolutional approaches), as the centroids are selected solely from the pooled response covariance statistics, which does not explicitly favor centered patches.

\section{Why Does Clustering Work?}
We will briefly discuss the reason why simple clustering algorithms work well in finding a good dictionary. Essentially, given a dictionary $\bD$ of size $K$ and specifications on the encoding and pooling operations, the feature extraction pipeline could be viewed as a projection from the original raw image pixels to the pooled feature space $\mathbb{R}^{K}$. Note that this space does not degenerate since the feature extraction is highly nonlinear, and one could also view this from a kernel perspective as defining a specific kernel between images. Then, one way to evaluate the information contained in this embedded feature space is to use the covariance matrix $\bC$ of the output features, which plays the dual role of the kernel matrix between the encoded patches. 

Considering that larger dictionaries almost always increases performance \cite{yang2010efficient}, and in the limit one could use all the patches $\mathcal{P}$ as the dictionary, leading to a $N$-dimensional space and an $N\times N$ covariance matrix $\bC_{\mathcal{P}}$. Since in practice we always assume a budget on the dictionary size, the goal is to find a dictionary $\mathcal{D}$ of size $K$, which yields a $K$-dimensional space and a covariance matrix $\bC_{\mathcal{D}}$. The approximation to the ``oracle'' encoded space using $\mathcal{D}$ could then be computed as:
\begin{equation}
    \bC_{\mathcal{P}} \approx \bC_{\mathcal{P}\mathcal{D}}\bC_{\mathcal{D}}^{+}\bC_{\mathcal{D}\mathcal{P}}
\end{equation}
where $\bC_{\mathcal{P}\mathcal{D}}$ is the covariance matrix between the features extracted by $\mathcal{P}$ and the one extracted by $\mathcal{D}$, and $\bC^{+}$ denotes the matrix pseudo-inverse. 

Note that such explanation works for both the before-pooling case and the after-pooling case. The dictionary learning algorithm can then be thought as finding a dictionary that approximates $\bC_{\mathcal{P}}$ best. Interestingly, this could be thought as a form of the \nystrom method that subsamples subsets of the matrix columns for approximation. The \nystrom method has been used to approximate large matrices for spectral clustering \cite{fowlkes2004spectral}, and here enables us to explain the mechanism of dictionary learning. Recent research in the machine learning field, notably \cite{kumar2012sampling}, supports the recent empirical observations in vision: first, it is known that uniformly sampling the columns of the matrix $\bC_{\mathcal{P}}$ already works well in reconstruction, which explains the good performance of random patches in feature learning \cite{saxe2011random}; second, theoretical results \cite{kumar2012sampling,zhang2008improved} have shown that clustering algorithms works particularly better than other methods as a data-driven way in finding good subsets to approximate the original matrix, justifying the use of clustering in the dictionary learning works. 

In the patch-based dictionary learning, clustering could be directly applied on the set of patches (\ie the largest dictionary in the limit). When we consider pooling, though, using all the patches becomes non-trivial, since we need to compute pooled features using each patch as a code, which is computationally overwhelming. Thus, a reasonable approach to consider is to first find a subset of all the patches using patch-based clustering (although this ``subset'' is still larger than our final dictionary size), and perform clustering on the pooled outputs of this subset only, which leads to the proposed algorithm. The performance of the algorithm is then supported by the \nystrom theory above.

Based on the matrix approximation explanation, we further reshape our selected features to match the distribution in the higher-dimensional space corresponding to the starting dictionary. After selecting $K$ centroids denoted by subset $\mathcal{S}$, we can group the original $M\times M$ covariance matrix as
\begin{equation}
    \bC = \begin{bmatrix}
        \bC_{\mathcal{S}\mathcal{S}} & \bC_{\mathcal{S}\bar{\mathcal{S}}} \\
        \bC_{\bar{\mathcal{S}}\mathcal{S}} & \bC_{\bar{\mathcal{S}}\bar{\mathcal{S}}}
        \end{bmatrix}, \quad%
    \bW = \begin{bmatrix} \bC_{\mathcal{S}\mathcal{S}} \\ \bC_{\bar{\mathcal{S}}\mathcal{S}} \end{bmatrix}
\end{equation}
where $\bC_{\mathcal{S}\bar{\mathcal{S}}}$ denotes the covariance between the subset $\mathcal{S}$ and the subset $\bar{\mathcal{S}}$ and so on. We then approximate the original high-dimensional covariance matrix as
\begin{equation}
    \bC \approx \bW\bC_{\mathcal{S}\mathcal{S}}^{+}\bW^\top
\end{equation}
More importantly, given the pooled outputs $\bx_{\mathcal{S}}$ using the selected filters, the high-dimensional feature could be approximated by 
\begin{equation}
\bx \approx \bA\bx_{\mathcal{S}}, \text{ where } \bA = \bW\bC_{\mathcal{S}\mathcal{S}}^{-1}
\end{equation}
Notice that the implicit dimensionality of the data is still no larger than the number $K$ of selected codes, so we can apply SVD on the matrix $\bA$ as $\bA = \bU\bLambda\bV$ where $\bLambda$ is a $K\times K$ diagonal matrix and $\bU$ is a $M\times K$ column-wise orthonormal matrix, and compute the low-dimensional feature (with a little abuse of terminology) as
\begin{equation}
    \bar{\bx}_{\mathcal{S}} = \bLambda\bV\bx_{\mathcal{S}}
\end{equation}
The $K\times K$ transform matrix $\bLambda\bV$ could be pre-computed during feature selection, and imposes minimum overhead during the actual feature extraction. This transform does not change the dimensionality or the rank of the data, and only changes the shape of the underlying data distribution. In practice, we found the transformed data yields a slightly better performance than the untransformed data when combined with a linear SVM with $L_2$ regularization.

If we simply would like to find a low-dimensional representation from the $M$-dimensional pooled features, one would naturally choose PCA to find the $K$ most significant projections:
\begin{equation}
    \bC \approx \bU_K\bLambda_K\bU_K^\top
\end{equation}
where the $M\times K$ matrix $U_K$ contains the eigenvectors and the $K\times K$ diagonal matrix $\bLambda_K$ contains the eigenvalues. The low-dimensional features are then computed as $\bx_K = \bU_K^\top\bx$.

We note that while this guarantees the best $K$-dimensional approximation, it does not help in our task since the number of filters are not reduced, as PCA almost always yields non-zero coefficients for all the dimensions. Linearly combining the codes does not work either, due to the nonlinear nature of the encoding algorithm. However, as we will show in Section \ref{subsec:exp:classification}, results with PCA show that a larger starting dictionary almost always help performance even when the feature is then reduced to a lower dimensional space of the same size as a smaller dictionary, which justifies the use of matrix approximation to explain the dictionary learning behavior.

\section{Experiments}
We apply our pooling-invariant dictionary learning (PDL) algorithm on several benchmark tasks, including the CIFAR-10 and STL datasets on which performance can be systematically analyzed, and the fine-grained classification task of classifying bird species, on which we show that feature learning provides a significant performance gain compared to conventional methods.

\begin{figure*}
    \centering
    \begin{tabular}{cccc}
        \includegraphics[height=0.2\textwidth]{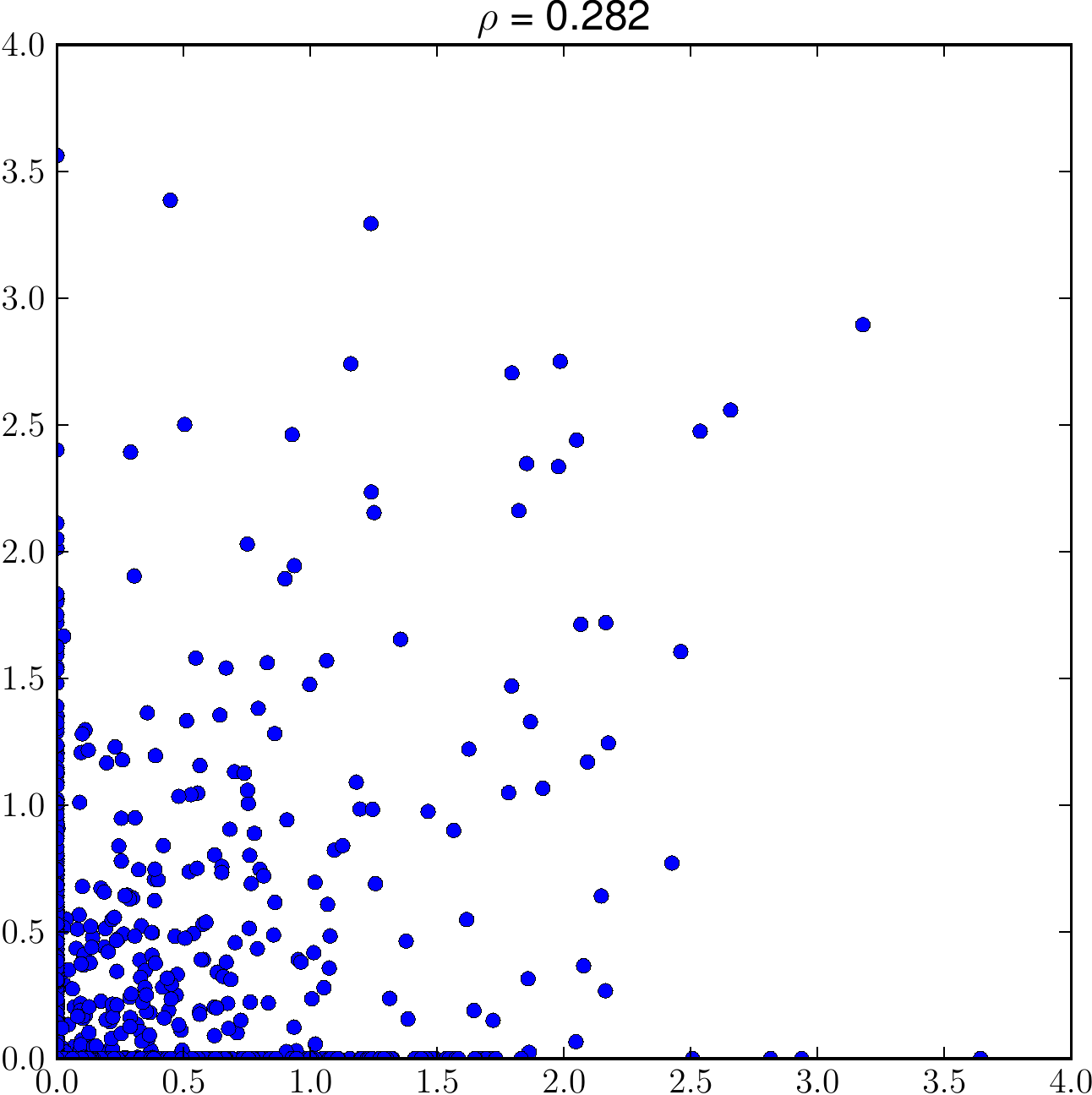} &
        \includegraphics[height=0.2\textwidth]{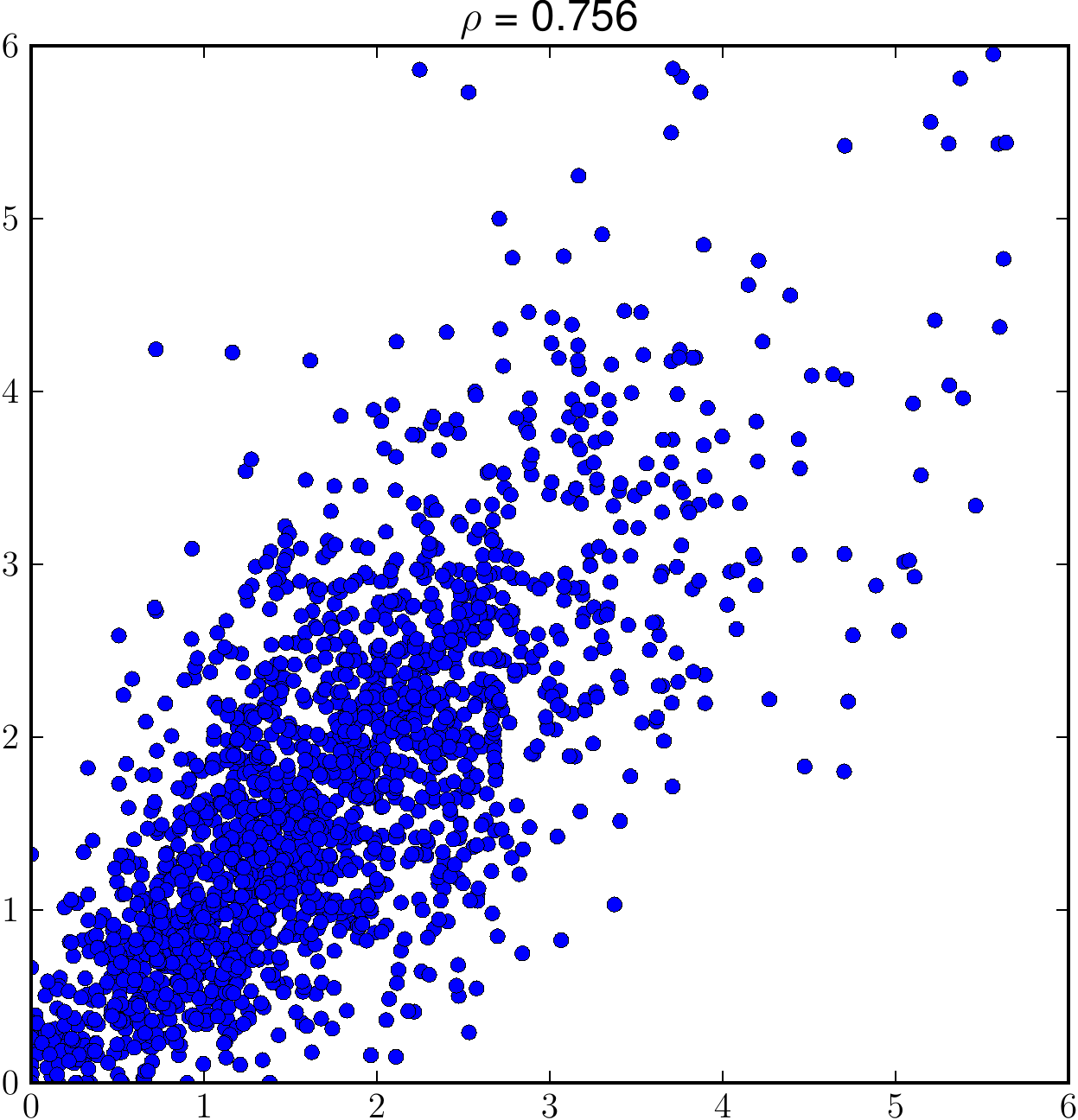} &
        \includegraphics[height=0.2\textwidth]{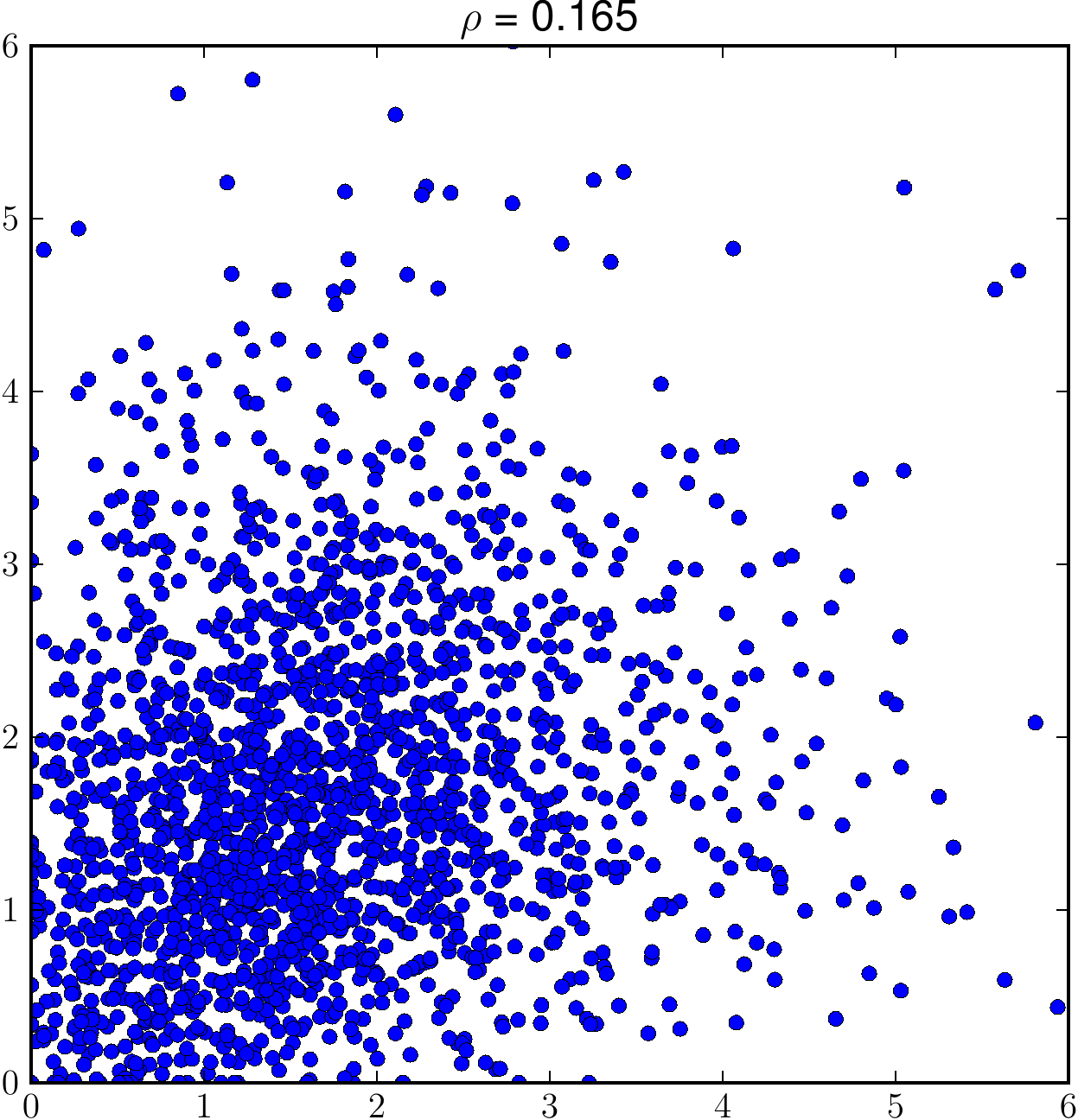} &
        \includegraphics[height=0.2\textwidth]{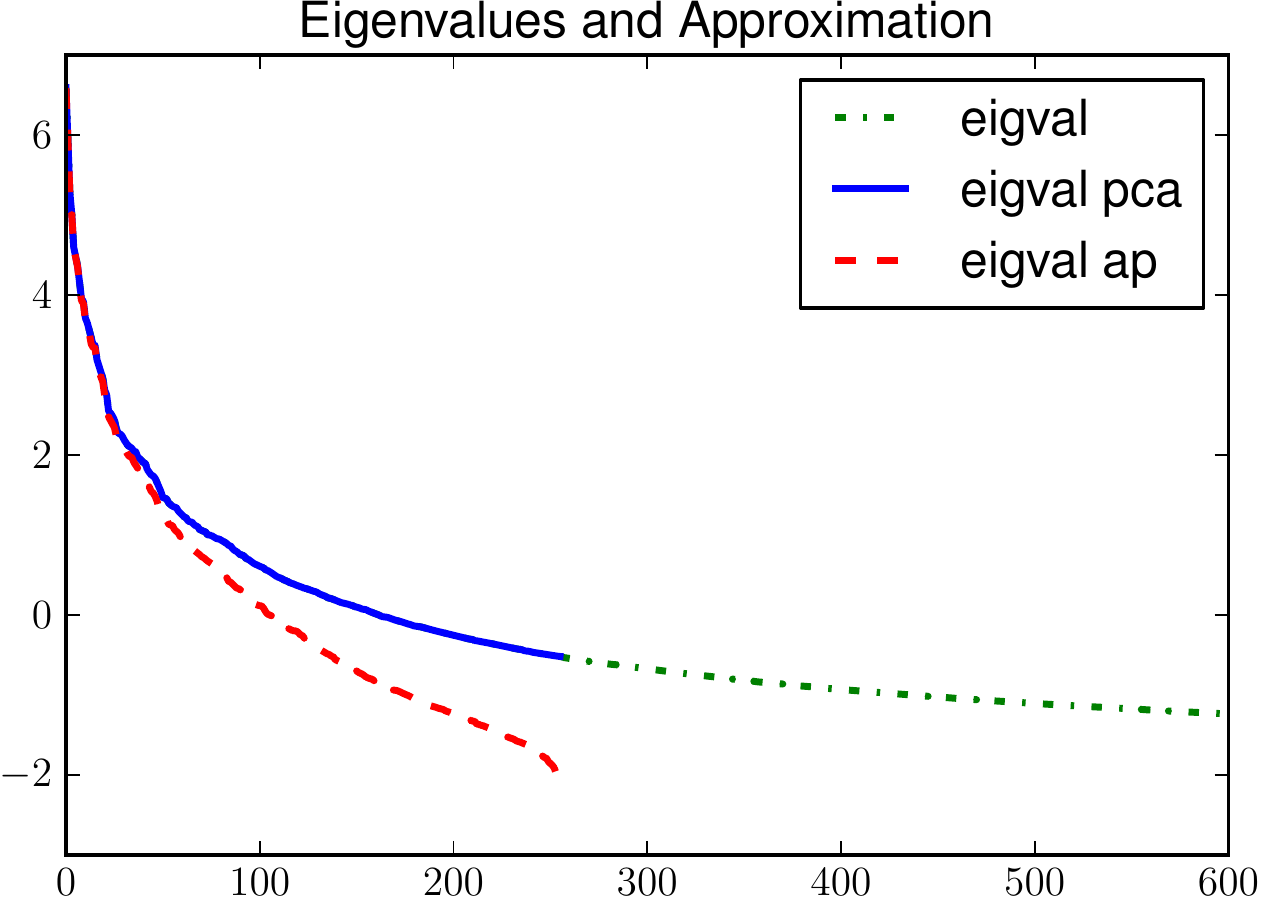}\\
        (a) & (b) & (c) & (d)
    \end{tabular}
    \caption{(a)-(c): The filter responses before and after pooling: (a) before pooling, between codes in the same cluster (correlation $\rho=0.282$), (b) after pooling, between codes in the same cluster ($\rho = 0.756$), and (c) after pooling, between the selected centroids ($\rho = 0.165$), (d): the approximation of the eigenvalues using the \nystrom method.}\label{fig:pairwiseresponses}
\end{figure*}

\subsection{CIFAR-10 and STL}
The CIFAR-10\footnote{\url{http://www.cs.toronto.edu/~kriz/cifar.html}} and the STL\footnote{\url{http://www.stanford.edu/~acoates/stl10/}, \cite{coates2010aistats}} datasets are extensively used to analyze the behavior of feature extraction pipelines. CIFAR-10 contains a large number of training and testing data, while STL contains a very small amount of training data and a large amount of unlabeled images. As our algorithm works with any encoding and pooling operations, we adopted the standard setting usually carried on the dataset: extracting local $6\times 6$ patches with mean subtracted and contrast normalized, whitening the patches with ZCA, and then train the dictionary with normalized K-means. The features are then encoded using one-sided threshold encoding with $\alpha=0.25$ and average pooled over the four quadrants ($2\times 2$) of the image. For STL, we followed \cite{coates2011selecting} and resized them to $32\times 32$. For the PDL algorithm, instead of learning a different set of codes for each pooling quadrant, we learn an identical set of codes in general for pooling the coded outputs. For all experiments, we carry out 5 independent runs and take the mean accuracy to report here.

\subsection{Statistics for Feature Selection}

We first verify whether the learned codes capture the pooling invariance as we claimed in the previous section. To this end, we start from the selected features in Section \ref{subsec:visualization}, and randomly sample three types of filter responses: (a) pairwise filter responses \emph{before pooling} between codes in the same cluster, (b) pairwise filter responses \emph{after pooling} between codes in the same cluster, and (c) pairwise filter responses after pooling between the selected centroids. The distribution of such responses are plotted in Figure \ref{fig:pairwiseresponses}. The result verifies our conjecture well: first, codes that produce uncorrelated responses before pooling may become correlated after the pooling stage (comparing \ref{fig:pairwiseresponses}(a) and \ref{fig:pairwiseresponses}(b)), which could effectively be identified by the affinity propagation algorithm; second, by explicitly taking into consideration the pooling behavior, we are able to select a subset of the features whose responses are lowly correlated (compare \ref{fig:pairwiseresponses}(b) and \ref{fig:pairwiseresponses}(c)), which helps preserve more information with a fixed number of codes.

Figure \ref{fig:pairwiseresponses}(d) shows the eigenvalues of the original covariance matrix and those of the approximated covariance matrix, using the same setting as in the previous subsection\footnote{Note that since we only select 256 features, the number of nonzero eigenvalues is 256 for the approximation.}. The approximation captures the largest eigenvalues of the original covariance matrix well, while dropping at a higher rate for smaller eigenvalues.

\begin{figure}
    \centering
    \includegraphics[width=0.38\textwidth]{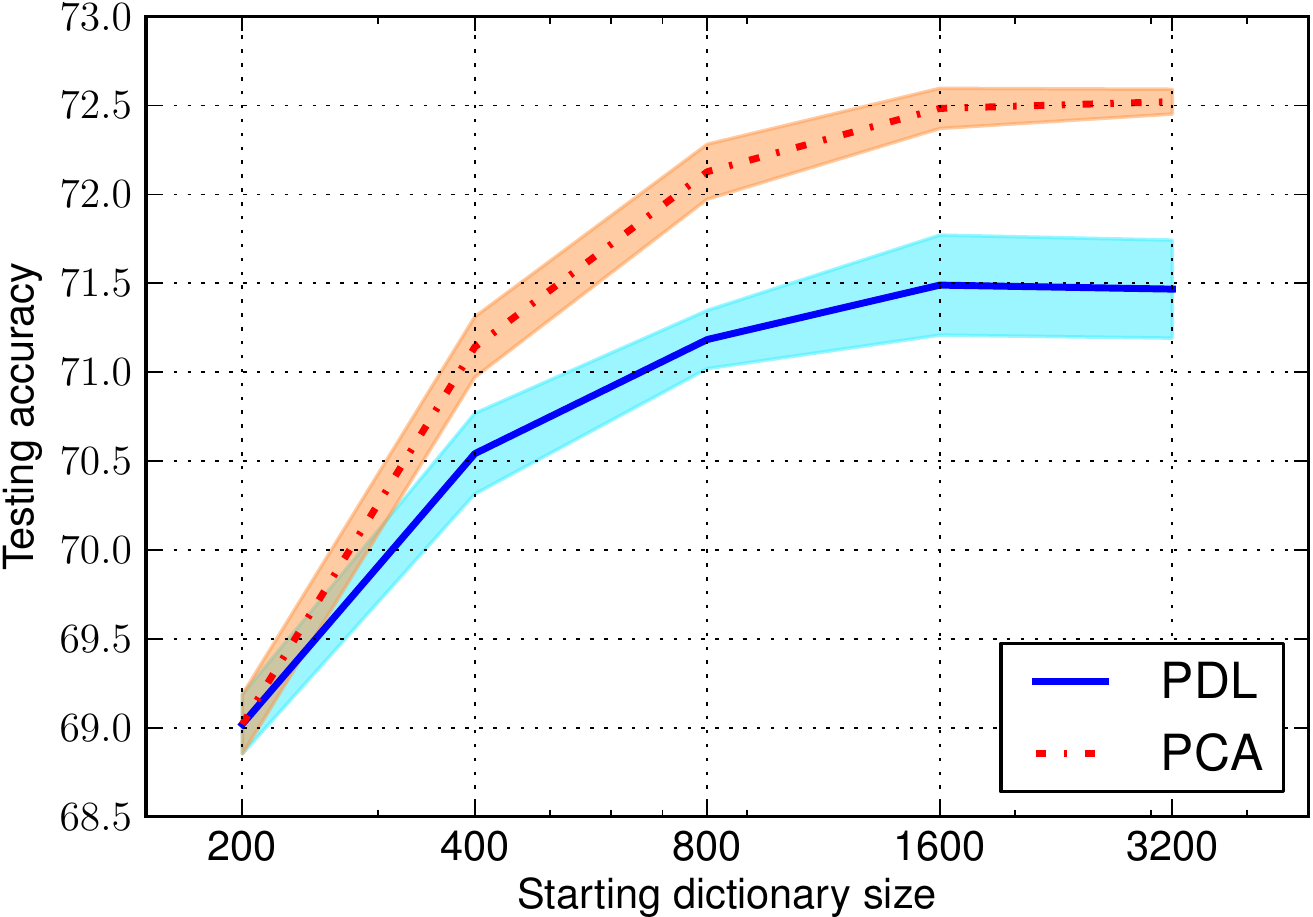}
    \caption{Performance improvement on CIFAR when using different starting dictionary sizes and a final dictionary of size 200. Shaded areas denote the standard deviation over different runs. Note that the x-axis is in log scale.}\label{fig:relativeimprovement}
\end{figure}

\subsection{Classification Performances}\label{subsec:exp:classification}
Figure \ref{fig:relativeimprovement} shows the relative improvement obtained on CIFAR-10, when we use a budgeted dictionary of size 200, but perform feature selection from a larger dictionary as indicated by the X axis. The PCA performance is also included in the figure as a loose upper bound of the feature selection performance. Learning the dictionary with our feature selection method consistently increases the performance as the size of the original dictionary increases, and is able to get about 70\% the performance gain as obtained by PCA (again, notice that PCA still requires all the codes to be used, thus does not save feature extraction time).

\begin{figure}
    \centering
    \includegraphics[width=0.35\textwidth]{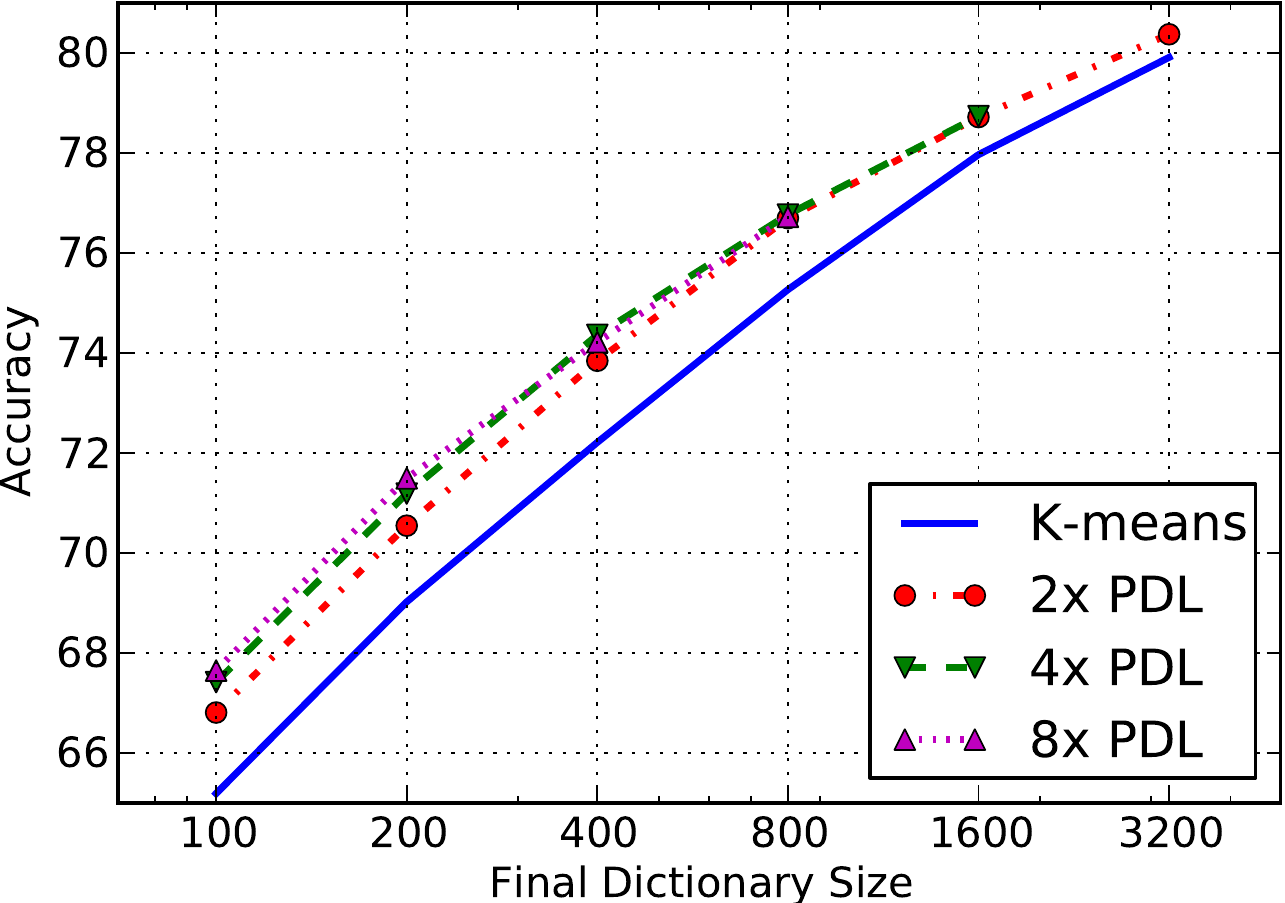}\\\vspace{0.1in}
    \includegraphics[width=0.35\textwidth]{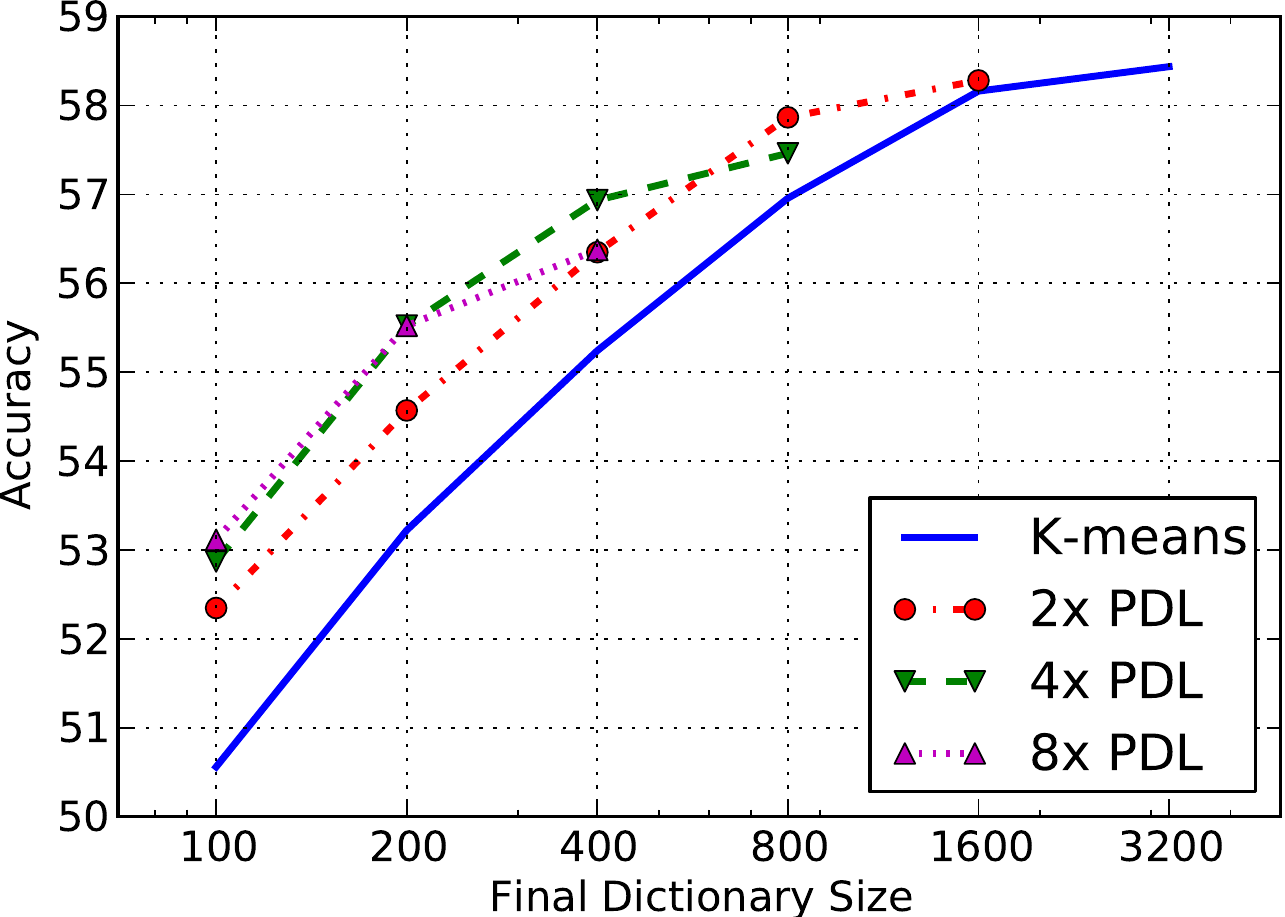}
    \caption{Accuracy values on the CIFAR-10 (top) and STL (bottom) datasets under different final dictionary size. ``nx PDL'' means learning the dictionary from a starting dictionary that is n times larger.}\label{fig:cifarstl}
    \vspace{0.3in}
    \includegraphics[width=0.23\textwidth]{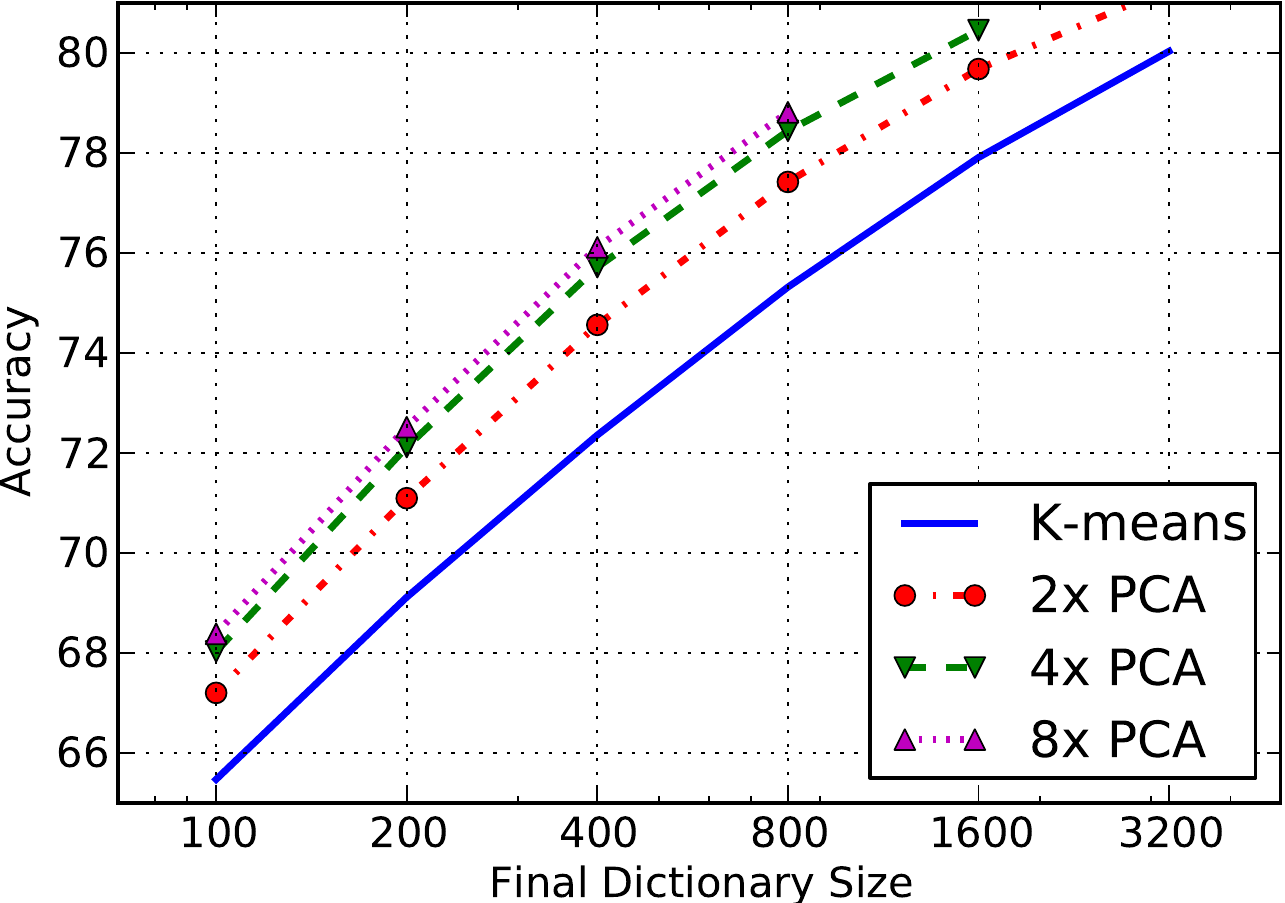}
    \includegraphics[width=0.23\textwidth]{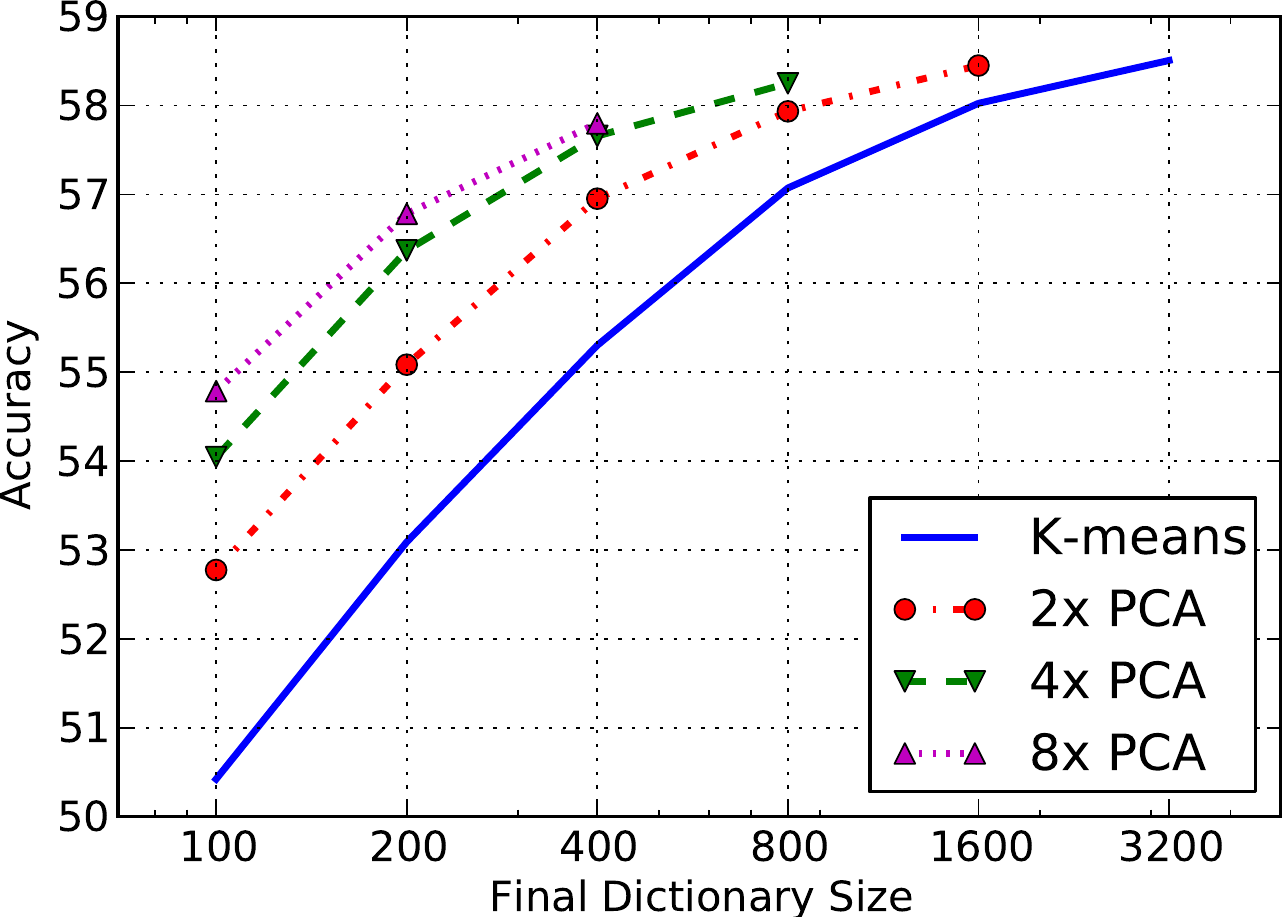}
    \caption{Accuracy values using PCA to reduce the dimensionality to the same as that PDL uses, as an accuracy upper bound for the performance of PDL.}\label{fig:cifarstl-svd}
\end{figure}

The detailed performance of our algorithm on the two datasets, using different starting and final dictionary sizes, is visualized in Figure \ref{fig:cifarstl}. Table \ref{tab:cifarstl} summarizes the accuracy values of two particular cases - final dictionary sizes of 200 and 1600 respectively. Note that our goal is not to get the best overall performance - as performance always goes up when we use more codes. Rather, we focus on how much gain the pooling-aware dictionary learning gets, given a fixed dictionary size as the budget. Figure \ref{fig:cifarstl-svd} shows the performance of the various settings, using PCA to reduce the dimensionality instead of PDL. As we stated in the previous section, this serves as an upper bound of the feature selection algorithms. 

Overall, considering the pooled feature statistics always help us to find better dictionaries, especially when only a small dictionary is allowed for classification. Choosing from a larger starting dictionary helps increasing performance, although such effect saturates when the dictionary size is much larger than the target size. For the STL dataset, a large starting dictionary may lessen the performance gain (Figure \ref{fig:cifarstl}(b)). We infer the reason to be that feature selection is more prone to local optimum, and the small training data of STL may cause the performance to be sensitive to suboptimal codebook choices. However, in general the codebook learned by PDL is consistently better than its patch-based counterpart. 

\begin{table}
    \centering
	\caption{Classification Accuracy on the CIFAR-10 and STL datasets under different budgets.}\label{tab:cifarstl}
    \begin{tabular}{c|c|c}
		\hline
		Task & Learning Method & Accuracy\\
		\hline
		            & K-means   & 69.02 \\
		CIFAR-10    & 2x PDL    & 70.54 (+1.52) \\
		200 codes   & 4x PDL    & 71.18 (+2.16) \\
		            & 8x PDL    & 71.49 (+2.47) \\
		\hline
		CIFAR-10    & K-means   & 77.97 \\
		1600 codes  & 2x PDL    & 78.71 (+0.74) \\
		\hline
		            & K-means   & 53.22 \\
		STL         & 2x PDL    & 54.57 (+1.35)\\
		200 codes   & 4x PDL    & 55.53 (+2.31)\\
		            & 8x PDL    & 55.52 (+2.30)\\
		\hline
		STL         & K-means   & 58.16 \\
		1600 codes  & 2x PDL    & 58.28 (+0.12) \\
		\hline      
	\end{tabular}
\end{table}

Finally, we note that due to the heavy-tailed nature of the encoded and pooled features (see the eigendecomposition of Figure \ref{fig:pairwiseresponses}), one can infer that the representations obtained with a budget would have a correspondingly bounded performance when combined with linear SVMs. In this paper we have focused on analyzing unsupervised approaches. Incorporating weakly supervised information to guide feature learning / selection or learning multiple layers of feature extraction would be particularly interesting, and would be a possible future direction.

\subsection{Fine-grained Classification}
To show the performance of the feature learning algorithm in the real-world image classification tasks, we tested the performance of our algorithm on the fine-grained classification task, using the 2011 Caltech-UCSD Birds dataset \cite{WelinderEtal2010}\footnote{\url{http://www.vision.caltech.edu/visipedia/CUB-200-2011.html}}. Classification of file-grained categories poses a significant challenge for the contemporary vision algorithms, as such classification tasks usually requires the identification of localized appearances like ``birds with yellow spotted feather on the belly'' which is hard to capture using manually designed features. Recent work on fine-grained classification usually focuses on the localization of parts \cite{farrell2011birdlets, zhang2012pose, duan2012discovering}, and still uses manually designed features. Yao et al. \cite{yao2012codebook} proposed to use a template-based approach for more powerful features, but the approach may be difficult to scale as the number of templates grow larger\footnote{Due to computation complexity, some early work such as \cite{farrell2011birdlets,yao2012codebook} do not scale up well, and only reported performance on subsets of the whole data [personal communication].}. In addition, due to the lack of training data in fine-grained classification tasks, whether supervised feature learning is useful or not is unclear yet.

We performed classification on all the 200 bird species provided. For the image pre-processing we followed the same setting as \cite{zhang2012pose, yao2012codebook} by cropping the images to be centered on the birds using 1.5$\times$ the size of the provided bounding boxes. We then resized each cropped image to $128\times128$ to avoid any artifact that may be introduced by the varying number of local features. The training data are expanded by simply mirroring each training image. Then, we extract $5\times 5$ local whitened patches as we did for CIFAR-10, encoded them using threshold encoding with a dictionary of size $1600$, and performed $4\times 4$ max pooling since the bird images are larger than those from CIFAR and STL. The dictionary is learned using our feature extraction pipeline, from an original set of $3200$ patch-based clustering centers. 

Our classification results, together with previous state-of-the-art baselines from \cite{zhang2012pose}, are reported in Table \ref{tab:birds}.
It is somewhat surprising to observe that feature learning provides a significant performance boost in this case, indicating that in addition to part localization (which has been the focus of fine-grained classification), learning appropriate features / descriptors to represent local appearances may be a major factor in fine-grained classification, possibly due to the subtle appearance changes for such tasks. As we have shown here, even simple and fully unsupervised feature learning algorithms such as K-means and PDL could lead to significant accuracy improvement, and we hope this would inspire further advancement in the fine-grained classification research.

\begin{table}
    \centering
	\caption{Classification Accuracy on the 2011 Caltech-UCSD Birds dataset.}\label{tab:birds}
    \begin{tabular}{c|c}
		\hline
		Method & Accuracy\\
		\hline
		BoW SIFT Baseline \cite{zhang2012pose} & 18.60 \\
		Pose Pooling + linear SVM \cite{zhang2012pose} & 24.21\\
		Pose Pooling + $\chi^2$ SVM \cite{zhang2012pose} & 28.18\\
		K-means + linear SVM (as in \cite{coates2011selecting}) & 38.17 \\
		PDL + linear SVM & {\bfseries 38.91}\\
		\hline
	\end{tabular}
\end{table}

\section{Conclusion}
We have proposed a novel algorithm to efficiently take into account the invariance of learned features after the spatial pooling stage. The algorithm is empirically shown to identify redundancy between codes learned in a patch-based way, and yields dictionaries that produces better classification accuracy than simple patch-based approaches. To explain the performance gain we proposed to take a matrix approximation view of the dictionary learning, and show the close connection between the proposed methods and the \nystrom method. The proposed method does not introduce overheads during classification time, and could be easily ``plugged in'' to the existing image classification pipelines.

{\footnotesize
\bibliographystyle{plain}
\bibliography{refs}
}

\end{document}